\documentclass[letterpaper, 10 pt, conference]{ieeeconf}  

\IEEEoverridecommandlockouts                              
\usepackage{cite}
\usepackage{amssymb}
\usepackage{amsmath}
\usepackage[table]{xcolor}
\usepackage{multirow}
\usepackage{graphicx}
\usepackage{xspace}
\usepackage{subcaption}
\usepackage{caption}
\usepackage{floatrow}

\usepackage[normalem]{ulem}

\newcommand{\cmap}{C-map\xspace}
\newcommand{\ours}{c2g-HOF\xspace}
\newcommand{\ctog}{cost-to-go\xspace}
\newcommand{\cspace}{C-space\xspace}
\newcommand{\dof}{DoF\xspace}
\newcommand{\ctognet}{c2g-network\xspace}

\title{Cost-to-Go Function Generating Networks for \\High Dimensional Motion Planning}
\author{Jinwook Huh, Volkan Isler, and Daniel D. Lee
\thanks{All authors are with the Samsung AI Center NY, 123 West 18th Street, New York, New York 10011}%
}
\date{October 2020}

\begin{document}

\maketitle


\begin{abstract}
This paper presents \ours{} networks which learn to generate cost-to-go functions for manipulator motion planning.  
%
The \ours architecture consists of a \ctog{} function over the configuration space represented as a neural network (\ctognet{}) as well as
a Higher Order Function (HOF) network  which outputs the weights of the \ctognet{} for a given input workspace. 
Both networks are trained end-to-end in a supervised fashion using costs computed from traditional motion planners. Once trained, \ours can generate a smooth and continuous \ctog{} function directly from workspace sensor inputs (represented as a point cloud in 3D or an image in 2D). At inference time, the weights of the \ctognet{} are computed very efficiently and near-optimal trajectories are generated by simply following the gradient of the \ctog{} function. 

We compare \ours{} with traditional planning algorithms for various robots and planning scenarios.  The experimental results indicate that planning with \ours{} is significantly faster than other motion planning algorithms, resulting in orders of magnitude improvement when including collision checking.  Furthermore, despite being trained from sparsely sampled trajectories in configuration space, \ours{} generalizes to generate smoother, and often lower cost, trajectories. We demonstrate \ctog{} based planning on a 7 \dof manipulator arm where motion planning in a complex workspace requires only 0.13 seconds for the entire trajectory.
 
\end{abstract}

\section{Introduction}

Motion planning is the problem of generating motion commands which take a robot from an initial configuration to a final configuration in a collision free manner. It is one of the most fundamental problems in robotics; however, efficiently planning motions for a high degree of freedom (\dof) robot in a complex workspace is still extremely challenging and has been studied extensively over the years~\cite{lavalle2006planning,lynch2017modern,choset2005principles}.

We present a new approach which uses traditional motion planning methods as supervision and learns to generate configuration space cost-to-go functions directly from the robot's workspace provided as input. Specifically, consider the \emph{cost-to-go} function $f_\mathcal{W}(s, t)$ which outputs the cost of going from a source configuration $s$ to a terminal configuration $t$ in a given workspace $\mathcal{W}$. Let $g(x) = f_\mathcal{W}(x, t)$ for a fixed $t$. The robot can arrive at $t$ by following the gradient of $g$ with respect to its inputs. We show that a \emph{Higher Order Function~(HOF)} network can directly generate $f_\mathcal{W}(s, t)$  from the workspace $\mathcal{W}$ for a specific manipulator it is trained on.

\begin{figure}[t]
    \centering
    \includegraphics[width=0.98\columnwidth]{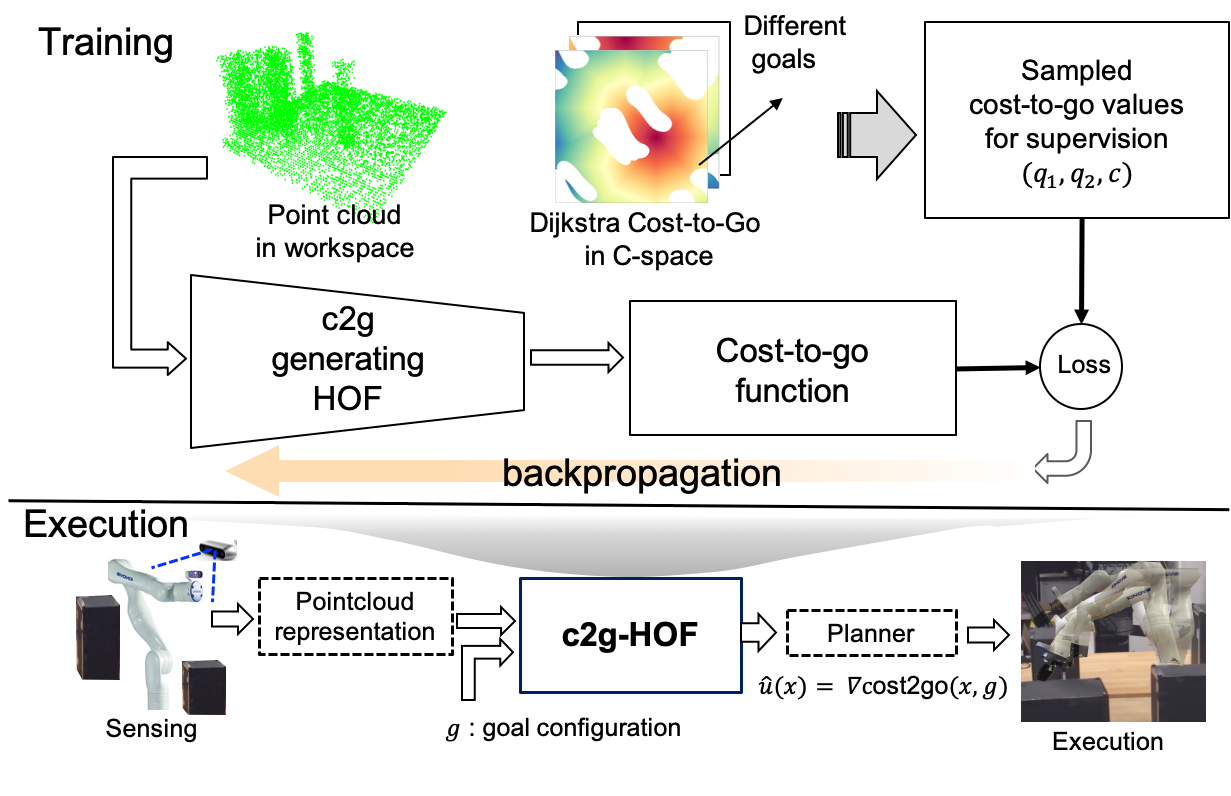}
    \caption{System Overview. 
    The \textbf{dataset} contains (1) randomly generated workspaces, and (2) for each workspace, the cost to go between pairs of sampled configurations.
    \textbf{Training architecture:} C2g generating HOF network encodes an input workspace and outputs the cost-to-go function represented as a radial basis function network. \textbf{Execution:} During run time,  following the gradient of the cost-to-go function yields continuous collision-free trajectories.   
    \label{fig:overview_flowchart}}
\vspace{-3mm}
\end{figure}

Fig. \ref{fig:overview_flowchart} presents an overview of our method, which we call \ours: The \ctog-generating HOF encodes an input workspace $\mathcal{W}$, and generates the weights of a radial basis function network (RBFN) which represents the \ctog{} function over the configuration space (\cspace). During training, the values generated by the \ctog{} function are compared to values generated by a traditional planner such as Dijkstra's algorithm over grid samples or a probabilistic roadmap over the configuration space. This way, the network learns to generate the \ctog{} function for any input workspace. 
During runtime, the weights are used to generate a compact neural network whose gradient yields the desired motion plan.

Our method \ours{} has a number of appealing properties. 
Its main advantage is that, once trained, it can instantly generate a continuous \ctog{} map over the entire C-map much faster than existing approaches. 
The \ctog{} function represented by a neural network (\ctognet{}) with weights from c2g generating HOF is compact and its gradient is readily computed to generate near-optimal trajectories.
Finally, we note that \ctognet yields a continuous, smooth function although it is trained from discretely sampled data.


In previous work~\cite{huh2020learning}, we investigated the problem of generating the \ctog{} function over the configuration space where we used an explicitly constructed configuration map (\cmap) as an intermediate step. In other words, for a $d$ \dof robot, the approach in~\cite{huh2020learning} created a $d$-dimensional binary map representing collisions which was then transformed into a \ctog{} map for a fixed destination. This approach worked well for low \dof robots where the \cmap can be represented explicitly on a grid. However, it does not scale well to high dimensions. In the present work, we overcome this difficulty by representing the \ctog function using a \emph{Radial Basis Function (RBF)} network. This allows the \ctog{} function to be generated directly from the workspace. The present work showcases further improvements regarding the network architecture: the \ctog{} is now learned for any target configuration rather than for a specific goal. Finally, we show how to implement various workspace representations including 2.5D workspaces as images and arbitrary 3D workspaces represented as dense point clouds encoded using PointNet \cite{qi2017pointnet}.


In summary, the core contributions of the paper are 1)~c2g-HOF architecture for generating continuous \ctog{} functions directly from the workspace. The approach eliminates the need for discretization, slow iterative propagation of cost values, and extensive collision checks; 2)~efficient trajectory planning based on \ctognet in \cspace for any target configuration; 3)~demonstration of the method's applicability to physical manipulators with high-dimensional complex C-spaces showing \ours{} results in prodigious improvements in motion planning efficiency.

\section{Related Work}
%
%
%

Due to its fundamental nature, there has been extensive prior research on motion planning for robotic manipulation~\cite{lavalle2006planning,lynch2017modern,choset2005principles,latombe1991robot,lavalle2001randomized, karaman2011anytime,kavraki1996probabilistic}. 
For low \dof robots, one can build a grid-based obstacle map in \cspace explicitly and use search-based planning algorithms such as Dijkstra or A* for robot motion planning. These search-based planning methods generate motions directly from the obstacle map based on sensor information~\cite{bonet2001planning,maximARA,stentz1995focussed}.
In potential field methods, local control laws are designed to steer away from obstacles and approach a goal point ~\cite{khatib1986real,rimon1992exact}. 
Since planning based on potential fields requires tuning the weights of potential field functions, previous work has proposed learning the weights from demonstration using gradient of cost functions \cite{ratliff2009learning,silver2008high}, apprenticeship learning \cite{abbeel2008apprenticeship}, Q-learning \cite{huh2018efficient}, and inverse optimal control \cite{finn2016guided}. Zeng et al. \cite{zeng2019end} suggest a deep neural network to learn a space-time cost volume based on trajectories from human demonstration and the trajectory with the minimum cost based on the learned cost volume is chosen for execution of a self-driving car. These previous methods are mostly focused on problems where obstacles and trajectories are represented in the same space, whereas \ours{} learns to generate near-optimal trajectories in a \cspace given obstacles sensed in a task space.


In high dimensional spaces, sampling based methods use random \cspace samples and collision checking in \cspace to build a graph-based representation of the \cmap. Probabilistic Roadmaps (PRMs) construct 
the graph by connecting nearby configurations and query a path between two vertices using local planners. Randomly exploring Random Trees (RRTs) explore collision-free \cspace by sampling robot's control inputs~\cite{lavalle2006planning,kavraki1996probabilistic,karaman2011anytime,bekris2003multiple}. It is known that the performance of these methods degrade significantly as the number of vertices and collision checks increase with \cspace dimensionality. 

Data-driven approaches have been investigated to improve the performance of traditional planners.
For example, some methods store previously constructed trees or trajectories and use them for new planning tasks when the workspace is similar to before \cite{hwang2015lazy,Phillips-RSS-12,berenson2012robot,coleman2015experience}. However, it is difficult to modify previously used planning structures when the environment is significantly changed.

Recently, several deep neural networks have been investigated for motion planning \cite{ichter2018learning,qureshi2018motion,molina2020learn, zhang2018learning}. Ichter et al.~\cite{ichter2018learning} learn sampling policies using a Conditional Variational Autoencoder (CVAE) for sampling-based planners. The work of \cite{qureshi2018motion} trains motion planning networks from demonstration by traditional planners and the networks output a trajectory directly given a workspace. Other approaches \cite{kumar2019lego,ichter2020learned} propose learning of critical points in C-space for efficient sampling-based planning. These neural network approaches commonly require hybrid approaches which combine neural networks with traditional planning algorithms \emph{for each instance} to find a trajectory whereas \ours{} requires supervision only for training.  It also generates a function over the entire configuration space whereas other neural network approaches output distributions of critical samples or trajectories directly. By generating \ctog{} functions over the entire C-space by HOF, we overcome sample complexity issues and directly generate efficient plans for various environmental conditions. We emphasize that the present architecture is a significant improvement from earlier work~\cite{huh2020learning} which required the collision map in the configuration space to first be constructed.

Reinforcement Learning (RL) approaches have also been investigated to utilize prior experience for robot planning and control \cite{peters2006policy,kober2009policy}. While these RL approaches focus on learning policies or state values for parameterized controllers from experience, our learning approach trains neural networks for the \ctog{} function in a supervised fashion.
Other work combining RL with planning methods \cite{francis2020long, faust2018prmrl,ekenna2015improved}. \cite{francis2020long,faust2018prmrl} overcome limitations of RL being trapped in local minima by utilizing global planners. \cite{ekenna2015improved} uses a local learning approach for PRM connections to adapt to spatial and temporal changes. 
\cite {zhang2019auto} suggests learning of networks from demonstration to generate a sequence of joint commands for a given task by combining mixture density models and long short-term memory network. Tamar et al.~\cite{tamar2016value} use a neural network to encode dynamic programming iterations for motion planning in a discrete space. In contrast, \ours{} directly learns a near-optimal, continuous \ctog{} function (\ctognet) as the weights of a radial basis function neural network.





\section{Problem Statement}

In this section, we formalize our problem setting and present an overview of our approach. 
\subsection{Problem Definition}

Let $f_{\mathcal{W}}: \mathcal{C}  \times \mathcal{C}  \rightarrow
[0, \infty)$
be the cost-to-go function which, for any given configurations $s, t$ in the configuration space $\mathcal{C}$ returns the cost to traverse a collision-free path from $s$ to $t$. 
The objective is to train a neural network which outputs $f_{\mathcal{W}}$ from input workspace $\mathcal{W}$.

Our method requires a representation of the workspace $\mathcal{W}$ as input. In this paper, we consider two such representations which cover a large number of application scenarios: (i)~3D obstacles represented as a point cloud, and (ii)~an overhead depth image of a 2.5D workspace. 

During the training phase, we assume that a planner which can compute the cost-to-go between two configurations is available. The specific metric used for computing the cost (e.g. length or energy) is determined by the planner.

\subsection{Method Overview}

\ours{} is trained in a supervised manner to closely mimic \ctog values given by the planner. Specifically, we use a uniform grid when possible or a Probabilistic Roadmap Method (PRM) to build a graph and use Diksjtra's method to compute shortest paths. The network is trained to output a \ctog function of two configurations in C-space. 



The network architecture used in this paper, \ours{}, consists of two components: (1)~the \ctog{} function represented as a neural network which can map any two configurations to a non-negative real number indicating the \ctog distance of these configurations (2)~HOF network which outputs the weights of the \ctog{} function from the workspace.

The HOF generating network requires an encoder for the workspace  $\mathcal{W}$ provided as input. 
We use
PointNet ~\cite{qi2017pointnet} for 3D point clouds. For 2D image or 3D voxel of workspace, we can apply image and voxel based convolution encoders. 
The \ctognet takes two configurations as input and outputs the optimal distance between two configurations.
To summarize, the c2g generating HOF takes the point cloud as input $\mathcal{W}$ and generates weights of a \ctog{} function $f_{\mathcal{W}} : \mathcal{C} \times \mathcal{C} \rightarrow [0, \infty)$, such that for all $q_1, q_2 \in \mathcal{C}$, 
$c = f_{\mathcal{W}}(q_1, q_2) \in [0, \infty)$ is the \ctog{} between $q_1$ and $q_2$. 



\section{Methods} \label{sec:methods}
In this section, we detail the architecture and the training procedure.


\subsection{Neural Network Architecture}

\begin{figure}[t]
\centering
\includegraphics[width=\columnwidth]{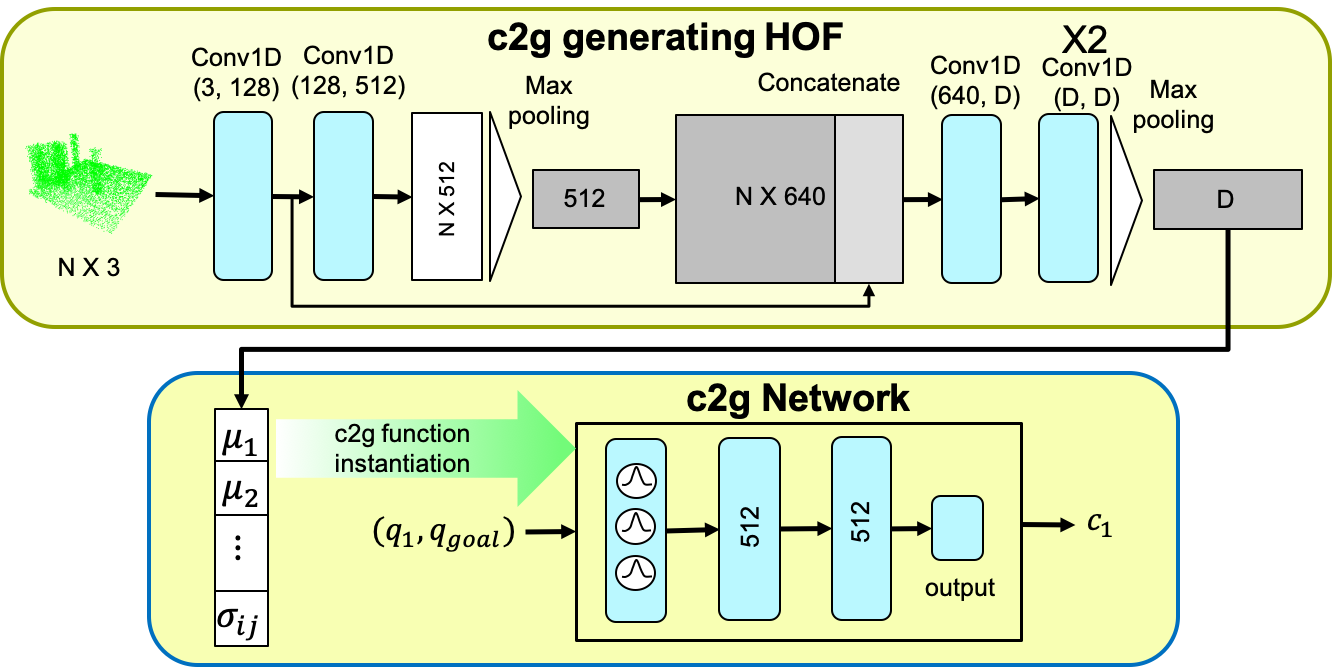}
\caption{\ours{} architecture. The workspace is represented as a point cloud. This $N \times 3$ ($N \times 2$ for two-dimensional representation of workspaces) point cloud is input to the c2g generating HOF network whose outputs are the weights of the \ctognet{}. These weights are transferred into a neural network (radial basis function network) which maps to \ctog{} values between robot configurations in a continuous fashion.}
\label{fig:total_network}
\end{figure}

The \ours{} architecture is composed of two subnetworks: c2g generating HOF and \ctog function network as shown in Fig. \ref{fig:total_network}. The head of the c2g generating HOF network is a PointNet encoder which has been shown to yield good performance in classification and segmentation of 3D point clouds. 


The encoding is then used to generate the weights of a RBFN with fully connected layers which yields the secondary subnetwork (\ctognet). We choose RBFN since radial basis functions are a good architecture to represent continuous \ctog{}. 
We tested and compared the performance with other networks such as fully connected networks. RBFN gave the best performance. 

Specifically, in our RBFN implementation, we use isotropic exponentials as basis functions whose output is followed by linear perceptrons: 
 We use 128 basis functions (64  for 2 \dof and 3 \dof manipulators) with two linear perceptrons at each layer followed by ReLU activation. Since the output of c2g generating HOF are reshaped into the parameters of \ctognet, the dimension of the output layer of c2g generating HOF should match the number of \ctognet parameters. In addition, the input dimension of \ctognet should match the dimension of input configurations. Since the \ctognet takes two input configurations, the input dimensions of 2 \dof, 3 \dof and 7 \dof should be 4, 6, and 14 for \ctognet. According to the input dimension, the dimensions of \ctognet and c2g generating HOF are adjusted accordingly.


\subsection{Dataset}

\begin{figure}[t]
\centering
\begin{subfigure}[b]{0.24\columnwidth}
\includegraphics[width=\textwidth]{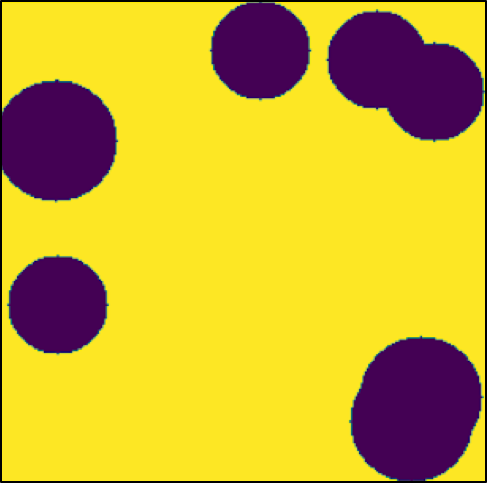}
\caption{Workspace} \label{fig:costmap_2d_workspace}
\end{subfigure}
\begin{subfigure}[b]{0.24\columnwidth}
\includegraphics[width=\textwidth]{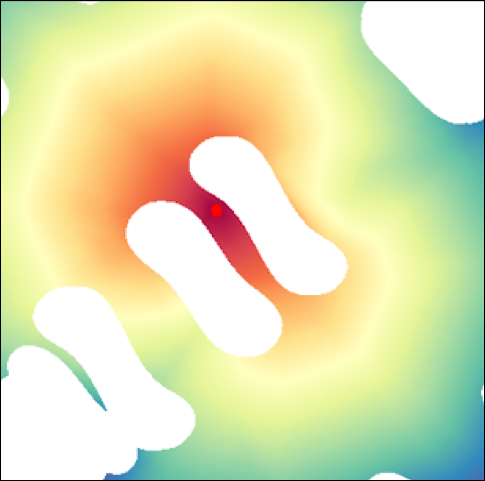}
\caption{Prediction} \label{fig:costmap_2d_pred}
\end{subfigure}
\begin{subfigure}[b]{0.24\columnwidth}
\includegraphics[width=\textwidth]{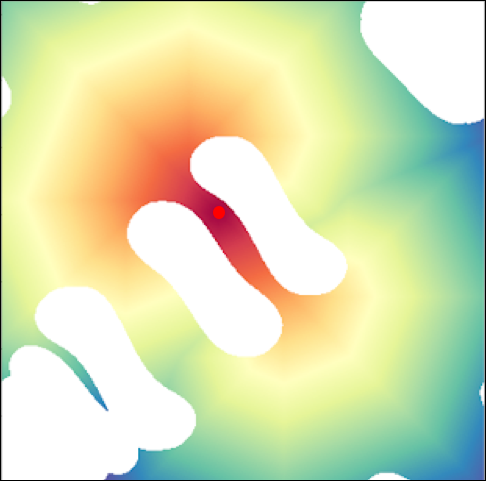}
\caption{Ground truth} \label{fig:costmap_2d_gt}
\end{subfigure}
\begin{subfigure}[b]{0.24\columnwidth}
\includegraphics[width=\textwidth]{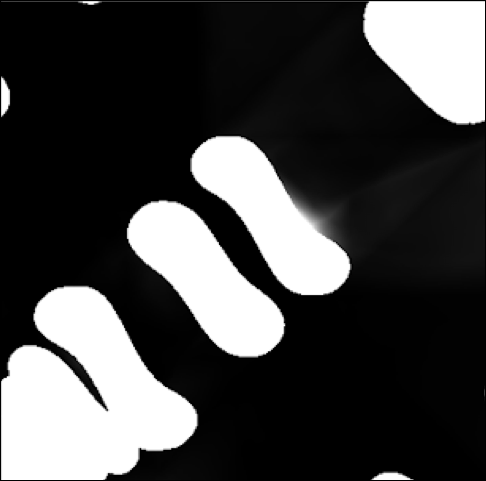}
\caption{Error map} \label{fig:costmap_2d_err}
\end{subfigure}
\caption{Cost map of test set in 2D \cspace.  We compute overall \ctog for a given goal and workspace for grid points over the \cspace  (Best viewed in color). We can see small errors in the boundary in the error map.}
\label{fig:costmap_2d}
\end{figure}

\begin{figure}
\centering
\begin{subfigure}[b]{0.35\columnwidth}
\includegraphics[width=\textwidth]{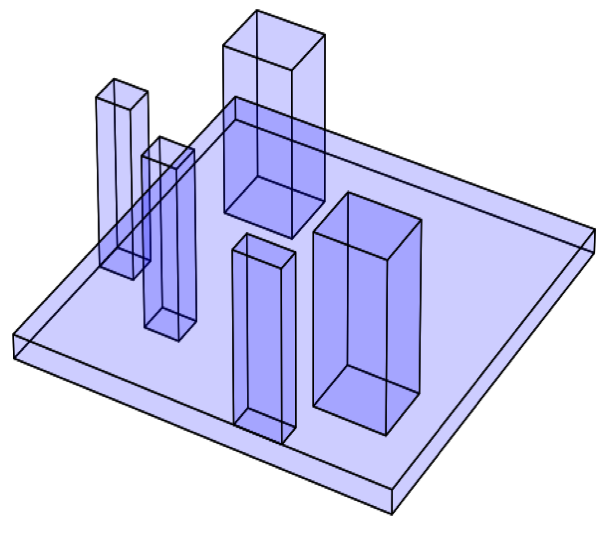}
\caption{Workspace} \label{fig:costmap_3d_workspace}
\end{subfigure}
\begin{subfigure}[b]{0.31\columnwidth}
\includegraphics[width=\textwidth]{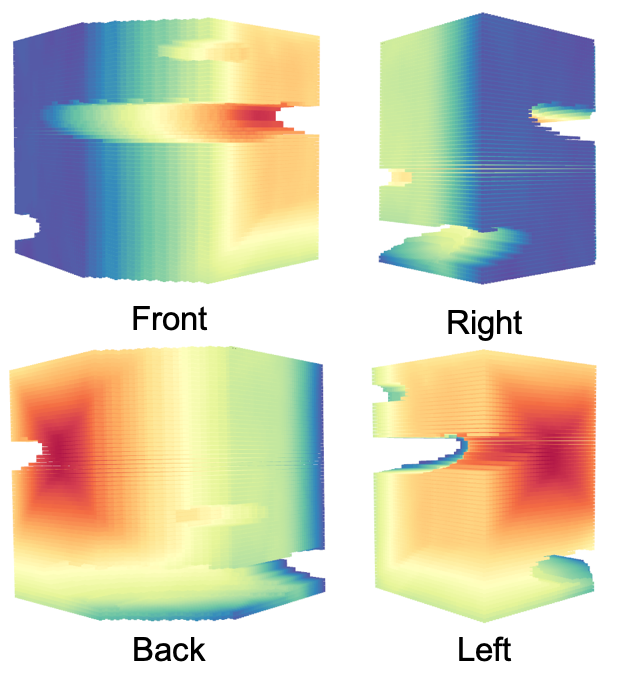}
\caption{Prediction} \label{fig:costmap_3d_pred}
\end{subfigure}
\begin{subfigure}[b]{0.31\columnwidth}
\includegraphics[width=\textwidth]{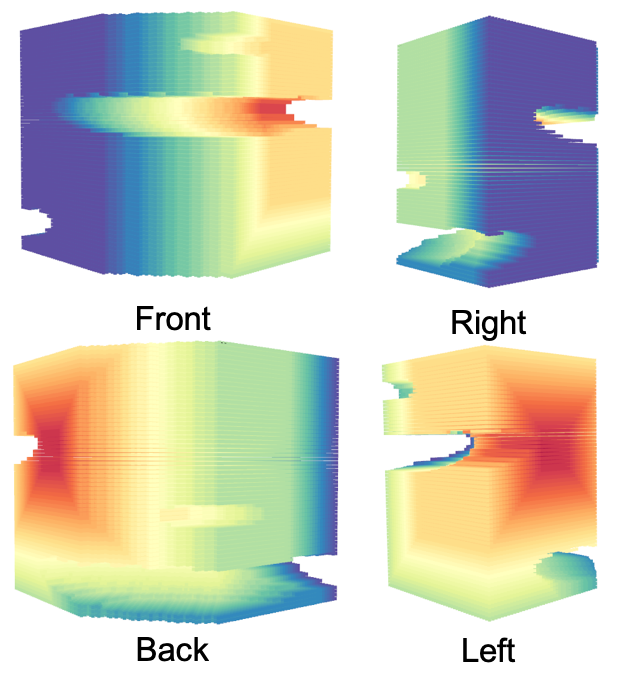}
\caption{Ground truth} \label{fig:costmap_3d_gt}
\end{subfigure}
\caption{Visualization of \ctog values of the 3 \dof robot in 3D workspace for a fixed goal over the entire \cspace. (Best viewed in color).}
\label{fig:costmap_3d}
\end{figure}

In order to train the network, we generated datasets for 2 \dof, 3 \dof, and 7 \dof manipulators separately. 
We set up a 2 link planar manipulator in 2D workspace with randomly generated circle obstacles from top-down view as shown in Figs. \ref{fig:costmap_2d_workspace} and \ref{fig:exp_setup_2d}. For the 3 \dof case, we set up a 3 \dof manipulator with two links (yaw and pitch joints at the robot base and one pitch joint between links) in 3D workspace with cuboid-shaped obstacles with random locations and dimensions on the table as shown in Figs. \ref{fig:costmap_3d_workspace} and \ref{fig:exp_setup_3d}. For 2 \dof and 3 \dof manipulators, the dataset is composed of 30,000 randomly generated workspaces (point clouds of obstacles).

To compute \ctog values in 2D and 3D \cspace we discretize the configuration space (360 $\times$ 360 cells in 2D \cspace and 64 $\times$ 64 $\times$  64 cells in 3D \cspace) and then compute \ctog of grids given a target goal point in the \cspace using Dijkstra's algorithm. Constructing connectivity between cells requires collision checks of cells in the entire configuration space. Note that from Dijkstra, we compute \ctog sets over the whole C-space for 500 sampled goal configurations. We compute these \ctog sets for each random workspace instance since \ctog must be computed separately for each new goal.
 
For 7 \dof, we generate 6,000 workspaces with random locations and dimensions of cuboid-shaped obstacles on the table. Uniform discretization becomes infeasible for 7 dimensional C-spaces. 
Instead, we construct Probabilistic Roadmaps (PRM) in \cspace and compute \ctog among the vertices by Dijkstra's algorithm. Specifically, the method generates 20,000 collision-free vertices, and then checks for connectivity between these vertices for PRM construction. 500 sets of cost-to-go values on the PRM vertices are then computed for randomly selected goal vertices. For the workspace representation, we use a PointNet encoding of the obstacle point cloud. A Kinova Gen3 manipulator arm is used for the 7 dof robot as shown in Fig. \ref{fig:exp_setup_7d}.



\subsection{Training}
The final dataset has \ctog sets composed of workspace point clouds and 2,500,000 \ctog tuples (paired by two configurations and \ctog value between configurations) corresponding to the point cloud. 
At each training iteration, we subsample the point cloud and randomly sample 100,000 \ctog tuples (2,000 in the 2 \dof case and 40,000 in the 3 \dof case) in the selected \ctog set.
The HOF generating network takes the point cloud as input. Its output weights are reshaped and used to construct the \ctognet. The sampled \ctog tuples are used for the prediction of \ctog by \ctognet. We chose Mean Square Error (MSE) between the predicted \ctog and ground-truth \ctog for the network loss function.
Since the output of c2g generating HOF network is used for the parameters of the \ctognet, the MSE error can be backpropagated to the c2g generating HOF network. This way, the network learns to generate parameters of the c2g-network which outputs the \ctog between two input configurations.
We train the network with the Adam optimizer with learning rate 3e-4 for 2,000 epochs on NVIDIA 1080 Ti GPUs.

\subsection{Trajectory Generation}
The robot trajectory to achieve the desired goal configuration is generated by following the gradient of the \ctog function starting from the initial configuration. In order to compute the gradient, we use the \ctog function as follows: we fix one configuration as the goal configuration, and perturb the other one in the direction of the gradient with respect to the current input configurations. The trajectory generated by following the \ctog gradient is then validated using a minimal number of collision checks.

\section{Results}

This section provides a series of experimental results which validate the proposed approach. We compare \ours{} with traditional approaches first in simulation and then on several physical robot arms.

\begin{table}[t]
\centering
\begin{tabular}{|c|c|c|c|}
\hline
\rowcolor[HTML]{FFFFC7} 
\textbf{\dof} & \textbf{RRT} & \textbf{$A^\star$}  & \textbf{c2g-HOF}       \\ \hline
2   & 2.789 (8.923)              & 0.934 (0.671)                & \textbf{0.168 (0.057)}              \\ \hline
3   & 1.792 (2.924) & 0.691 (0.516) &  \textbf{0.0729 (0.070)} \\ \hline \hline
\rowcolor[HTML]{FFFFC7} 
\textbf{\dof} & \textbf{RRT} & \textbf{RRT-smooth} & \textbf{c2g-HOF}       \\ \hline
7   & 1.044 (3.984)              & 1.8999 (4.097)                &  \textbf{0.135 (0.103)}             \\ \hline
\end{tabular}
\caption{\label{tab:table_time} Mean and standard deviation of planning times. In all cases, \ours is faster than RRT and $A^\star$.}
\vspace{-3mm}
\end{table}

\begin{table}
\centering
\begin{tabular}{|c|c|c|c|}
\hline
\rowcolor[HTML]{FFFFC7} 
\textbf{\dof} & \textbf{RRT} & \textbf{$A^\star$}  & \textbf{c2g-HOF}       \\ \hline
2   & 1.331 (0.425) & 1.0 (0.0.300)    & \textbf{0.974 (0.305)} \\ \hline
3   & 1.369 (0.267) & 1.0 (0.180)& \textbf{1.219 (0.211)} \\ \hline \hline
\rowcolor[HTML]{FFFFC7} 
\textbf{\dof} & \textbf{RRT} & \textbf{RRT-smooth} & \textbf{c2g-HOF}       \\ \hline
7   & 1.931 (0.608)  & 1.0 (0.237)   & \textbf{1.4027 (0.4375)} \\ \hline
\end{tabular}
\caption{\label{tab:table_length} Mean and standard deviation of trajectory length (The trajectory length is normalized by the length of the $A^\star$ value for 2 and 3 \dof; and by the length of RRT-smooth for 7 \dof). The \ours has shorter distance compared to RRTs.}
\vspace{-3mm}
\end{table}

\subsection{Learning of \ctog{}}
 
We measured the performance of the network in complex  environments. Fig. \ref{fig:costmap_2d} and \ref{fig:costmap_3d} show comparison of \ours{} after training against the optimal solution given by Dijkstra's algorithm on grids for 2D and 3D on novel test instances (unseen during training). The cost map for grid points in \cspace is computed with a given target configuration (red dot) and workspace. Note that \ours{} generates  smooth and continuous \ctog values over the entire \cspace although it was trained with a discretized dataset. In Fig. \ref{fig:costmap_2d_err}, we see that only small errors are present around the boundary of collision regions.

\subsection{Evaluation}
\label{sec:simulation_exp}

\begin{figure}[t]
\centering
\begin{subfigure}[b]{0.47\columnwidth}
\includegraphics[width=\textwidth]{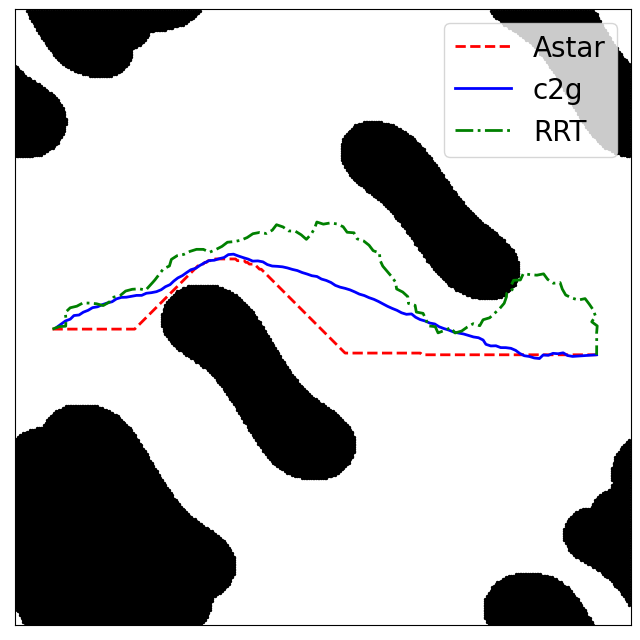}
\caption{Case 1} \label{fig:pathplanning2d_case1}
\end{subfigure}
\begin{subfigure}[b]{0.47\columnwidth}
\includegraphics[width=\textwidth]{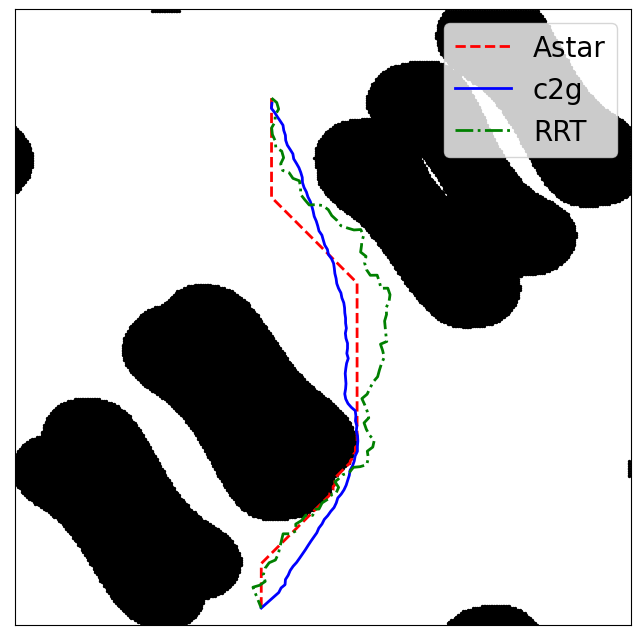}
\caption{Case 2} \label{fig:pathplanning2d_case2}
\end{subfigure}
\caption{Path planning result in a 2 \dof case.  \ours{} yields shorter than the $A^\star$ trajectory length even though $A^\star$ is optimal for the given discretization level.}
\label{fig:pathplanning2d}
\vspace{-3mm}
\end{figure}




For the evaluation of \ours{}, we focus on its performance in novel test environments and compare it against  conventional planners.
We set up three sets of environments with a 2-link manipulator in 2D workspace, a 3 \dof manipulator in 3D workspace, and a 7 \dof manipulator in 3D workspace. 

For each setting, we generate obstacles with random locations and sizes in workspace and then evaluate performance for 10,000 trajectories (200 workplaces $\times$ 50 random trajectories in each workspace) separately. We apply rules to ensure that the start and target configurations are sufficiently far apart ($180^{\circ}$ in 2 \dof and 3 \dof, and $360^{\circ}$ in 7 \dof). In 2D and 3D setups, we compare \ours with RRT and $A^\star$ in terms of the length of trajectory and planning time. In the 7D setup, we evaluate \ours against trajectories and planning time by RRT and RRT-smooth which performs post-processing to minimize the length of trajectory.

Table \ref{tab:table_time} and \ref{tab:table_length} show comparison results of 2 \dof, 3\dof, and 7 \dof between \ours{}, RRT, $A^\star$, and RRT-smooth in terms of planning time and trajectory length. The trajectory length is normalized by the length of the $A^\star$ (2 \dof and 3 \dof) and RRT-smooth (7 \dof) in the same start and goal trajectory. 
These baseline implementations are based on~\cite{kuffner2000rrt,russell2002artificial}.

In the 2 \dof case, the average planning of \ours{} for 10,000 trajectories is 16 times faster than RRT, and 5 times faster than $A^\star$ with a smaller variance. 
The \ours{} generates a trajectory based on the gradient of \ctog{} without space discretization or extensive collision checks. The running time does not vary significantly across instances as shown by the low variance of the planning time of \ours{}. In addition, the average trajectory length with \ours{} is 36\% shorter than the length of RRT and 3\% shorter than $A^\star$ trajectory length. The reason that \ours{} is shorter than the $A^\star$ trajectory is that \ours{} plans a continuous trajectory whereas $A^\star$ is limited by the grid discretization. As a result, $A^\star$ results in longer distances than \ours{} as shown in Fig. \ref{fig:pathplanning2d}.
 
In the 3 \dof case, \ours{} is 9 times faster than $A^\star$ and 24 times faster than RRT. In this scenario, \ours{} generates 22\% longer trajectories than the optimal $A^\star$ trajectory and 37\% shorter distance than RRT. Since the workspace for the 3 \dof case is relatively less cluttered than the workspaces of 2 \dof and 7 \dof, the overall planning time of all planners are smaller.


In the 7 \dof case, RRT has bi-directional tree structures from start and goal configurations and it has 3\% goal biased sampling to make the trajectory generation fast. Since $A^\star$ is intractable in 7 \dof, RRT-smooth with a path smoothing process after planning by RRT is used for comparison of trajectory quality. We apply the shortcut method \cite{Geraerts01082007} for path smoothing in this paper.

Overall, \ours{} has 7 times faster planning and 38\% shorter trajectory length than RRT. Although \ours{} is 40\% longer than the length of the RRT-smooth trajectory, RRT-smooth requires 14 times longer computation time for generating smooth paths. The RRT is efficient for trajectories in mostly open conditions, but \ours{} exhibits significant improvement for complicated trajectory generation tasks since the RRT planning and the path smoothing are inefficient in complicated environments conditions as shown in Section \ref{sec:physical_exp}.

 
In these experiments, we can see that \ours{} can learn to effectively replicate the \ctog{} by the given planners. It also has better performance in terms of planning time due to its fast neural network framework. Moreover, the continuous nature of the \ctognet{} output directly generates smooth motion plans in complex environments.


\begin{figure}[t]
\centering
\begin{subfigure}[b]{0.31\columnwidth}
\includegraphics[width=\textwidth]{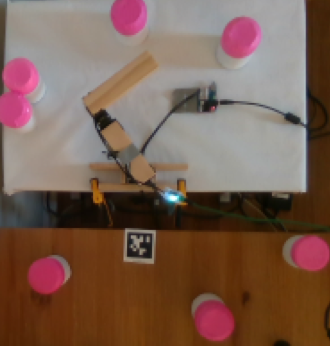}
\caption{2-link} \label{fig:exp_setup_2d}
\end{subfigure}
\begin{subfigure}[b]{0.31\columnwidth}
\includegraphics[width=\textwidth]{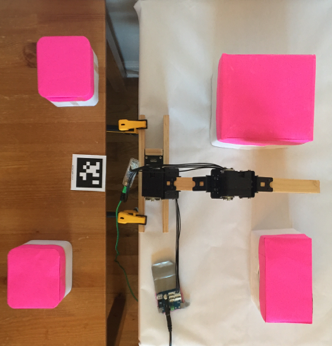}
\caption{3 \dof} \label{fig:exp_setup_3d}
\end{subfigure}
\begin{subfigure}[b]{0.31\columnwidth}
\includegraphics[width=\textwidth]{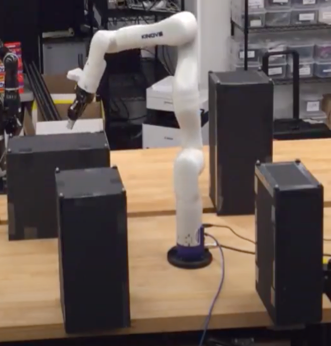}
\caption{7 \dof} \label{fig:exp_setup_7d}
\end{subfigure}
\caption{Physical robots used in experiments. We tested \ours in various configuration spaces including those with narrow passages.}
\label{fig:exp_setup}
\vspace{-3mm}
\end{figure}

\subsection{Physical Robot Experiments}
\label{sec:physical_exp}

\subsubsection{Experimental setup}
To verify the performance of \ours{} on hardware robots, we conducted experiments with real 2, 3, and 7 \dof manipulators. In the 2 \dof setup, the workspace is represented directly from an overhead Realsense D435 camera (Fig.~\ref{fig:exp_setup_2d}). From an input image, it detects the color of obstacles and randomly samples points from obstacle regions in the image for the input of c2g generating HOF.

In the 3 \dof setup, we use the same camera to detect obstacles in the workspace. Since we assume that the height of obstacles is constant, we can obtain point clouds from the overhead image (Fig.~\ref{fig:exp_setup_3d}). The robot has joint limits in the second and third joints, and it has same kinematics as the model trained in simulation.

In the 7 \dof setup, we use a Vicon system in order to cover whole workspace. The dimensions and locations of obstacles are obtained by Vicon. From four markers on the top of each obstacle, we construct a cubicle-shape object whose point cloud is randomly generated from obstacle regions. The obstacles are randomly arranged with the presence of narrow passages as shown in Fig. \ref{fig:exp_setup_7d}. We use Kinova Gen3 for the 7 \dof robot.

\begin{figure}[t]
\centering
\includegraphics[width=\columnwidth]{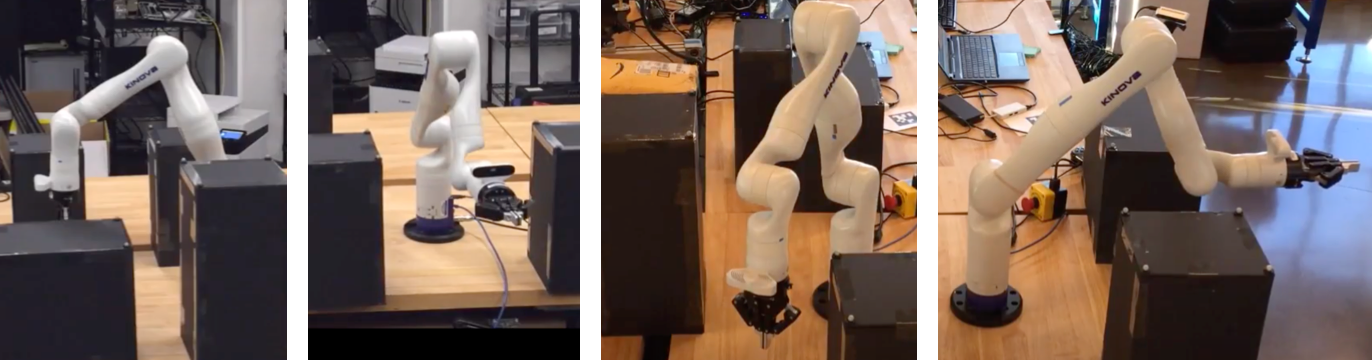}
\caption{Examples of end effector configurations for the 7 \dof manipulator. The target configurations are located in narrow passageways so that reaching them requires flipping end effector joints and rotating the base joint over obstacles. 
}
\label{fig:eef_configurations}
\vspace{-3mm}
\end{figure}

Target configurations are randomly generated and are intentionally placed in the narrow passages. The target configurations are very different from start configurations and require flipping end effector joints (5th and 6th joints) and rotating the base joint (1st joint) over obstacles. The start and goal configurations are chosen to point towards obstacles resulting in convoluted optimal trajectories as shown in Fig.~\ref{fig:eef_configurations}.


Each scenario has 10 target configurations and planners generate a new trajectory with the next target goal as soon as the manipulator arrives a target configuration. We only consider planning time for the comparison of trajectory generation time. We tested \ours and other approaches with the same target configurations in the same environment.

We verify \ours{} by generating the trajectory by following the gradient of \ctog{} values and see that the robot performs successfully even in the presence of narrow passages such as inserting the link between obstacles while avoiding collisions. For comparison, we test the performance of RRT and RRT-smooth using the same parameters as in Section \ref{sec:simulation_exp}. In addition, we also include PRM and PRM-Smooth results since we trained the network with PRMs. For PRM, we generate a roadmap with 8,000 vertices including start and goal configurations, and then update the roadmap by removing invalid nodes whenever the environment is changed and then checking the feasibility of final path. If the final path is not feasible, we remove the infeasible edges and replan  until a feasible path is found.

\begin{figure}[t]
\centering
\begin{subfigure}[b]{0.47\columnwidth}
\includegraphics[width=\textwidth]{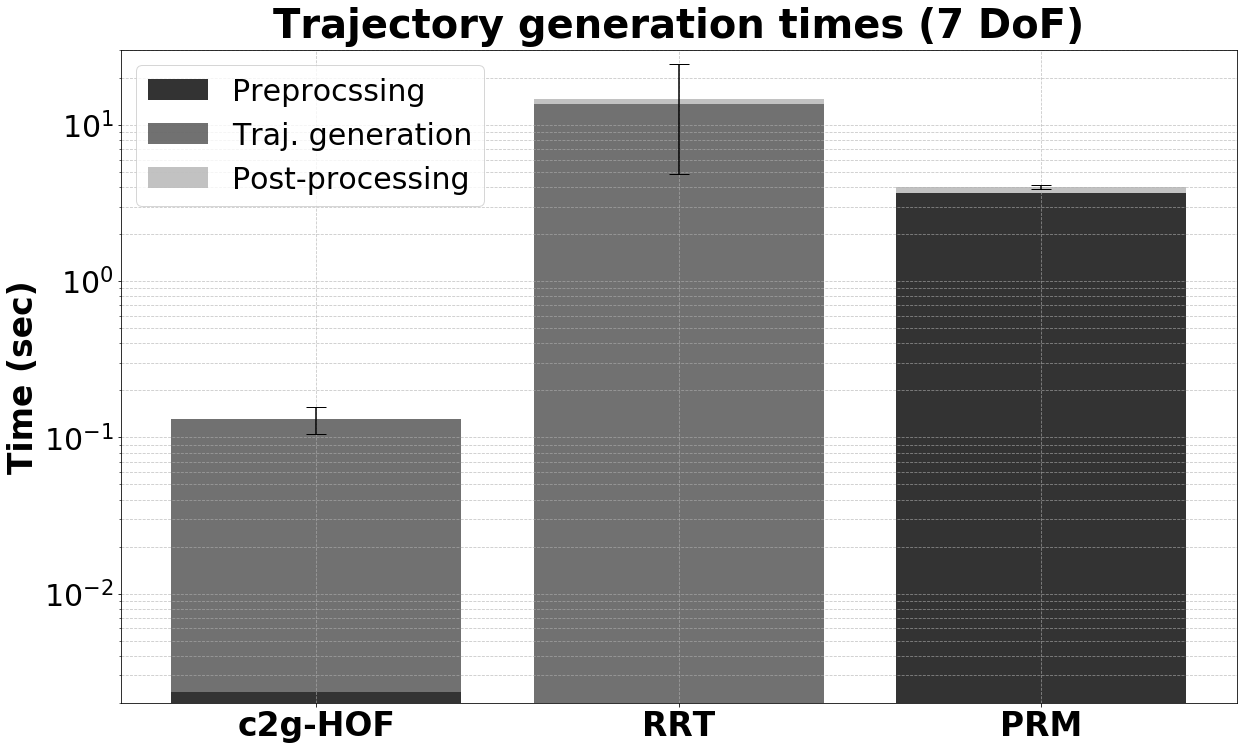}
\caption{Time} \label{fig:result_gen3_time}
\end{subfigure}
\begin{subfigure}[b]{0.47\columnwidth}
\includegraphics[width=\textwidth]{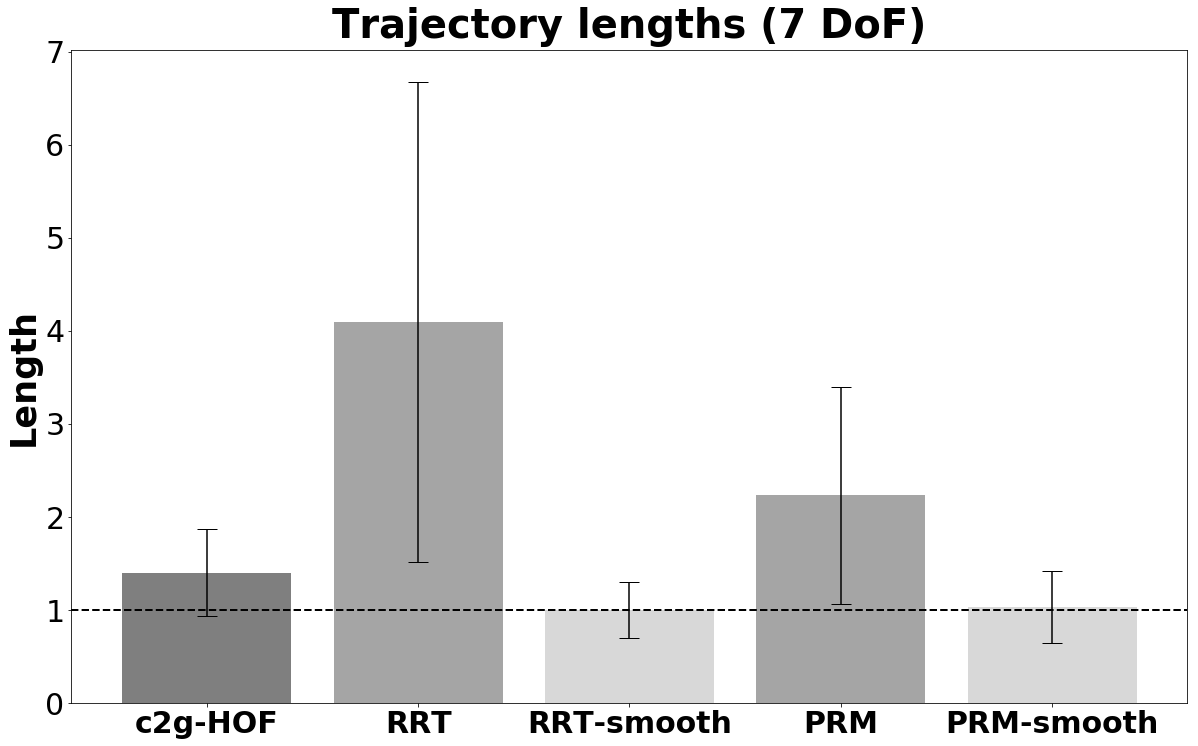}
\caption{Length} \label{fig:result_gen3_length}
\end{subfigure}
\caption{Trajectory results with the physical 7 \dof robot. \ours shows significant performance improvement in planning across a wide range of start and target configurations which require going through  narrow passages. }
\label{fig:result_gen3}
\vspace{-4mm}
\end{figure}

Fig. \ref{fig:result_gen3} and Table. \ref{tab:table_physical_time} show comparison results with the 7 \dof manipulator. We average planning times and trajectory lengths of 30 trajectories (10 trajectories $\times$ three workspaces). We see that \ours{} show increased relative performance over the simulation results since tasks are composed of more complicated target configurations than the randomly sampled simulation configurations. An advantage of \ours{} is that its performance does not degrade much as the environment complexity increases as compared to other planners.

\begin{table}[]
\begin{tabular}{|c|c|c|c|c|}
\hline
\rowcolor[HTML]{FFFFC7} 
\textbf{}    & \textbf{\begin{tabular}[c]{@{}c@{}}Preproc\end{tabular}} & \textbf{\begin{tabular}[c]{@{}c@{}}Traj\\ generation\end{tabular}} & \textbf{\begin{tabular}[c]{@{}c@{}}Postproc\end{tabular}} & \textbf{\begin{tabular}[c]{@{}c@{}}Total\\ time\end{tabular}} \\ \hline
\textbf{\ours} & 0.0023 & 0.129   & - & \textbf{0.131} \\ \hline
\textbf{RRT} & -  & 13.577 & 1.201  & 14.778 \\ \hline
\textbf{PRM} & 3.647 & 0.025 & 0.354 & 4.026 \\ \hline
\end{tabular}
\vspace{-3mm}
\caption{\label{tab:table_physical_time} Computation time breakdown. 
Preprocessing time for \ours is the time to generate the weights of the cost-to-go function by running the HOF network. There is no need for post-processing after trajectory generation. RRT does not require preprocessing. For PRM, preprocessing time is the time to update the roadmap. 
Both RRT and PRM require post-processing time to smooth the final  path. \ours is up to two orders of magnitude faster on the average.
}
\end{table}

Overall, these results establish \ours{} as an efficient and practical method which can learn from existing motion planners and generalize to generate continuous near-optimal cost-to-go functions in novel environments.

\section{Conclusion}
\vspace{-0.7mm}

This paper presented \ours{} for generating \ctog functions for
high degree of freedom manipulators. 
Learning the cost-to-go function in high-dimensional configuration spaces requires a large number of samples which, in turn, makes learning difficult. We overcome this challenge by introducing a novel Higher Order Function (HOF) architecture which learns to generate a continuous \ctog{}  function. A radial basis network is used to represent the \ctog{} function whose weights are generated by the higher order function network for arbitrary input workspaces.


Once the networks are trained, c2g generating HOF can almost instantly output the parameters of the \ctognet for a given obstacle point cloud. The \ctognet can then be used for planning effectively simply by following the gradient of \ctog.
Our experiments demonstrate that \ours{} exhibits significant performance improvement over conventional approaches for several kinds of manipulators, including a physical 7 \dof manipulator. On the average, planning with \ours{} is 14 times faster than RRTs in 7-dimensional space and 5 times faster than $A^\star$ in low dimensional spaces. Even in the presence of narrow passages for a 7 \dof manipulator, \ours{} is able to generate a feasible, near-optimal motion trajectory in only 0.13 seconds.

We are currently working on releasing our dataset and code. Directions for downloading them will be included in a later version of the paper. There are many exciting avenues for future work. We plan to incorporate the \ours{} framework into task driven planners. We also would like to extend the architecture for more dynamic planning scenarios and to accept partial environment information.

\bibliographystyle{IEEEtran}
\bibliography{references}


\end{document}